\documentclass[preprint,12pt,3p]{elsarticle}

\usepackage{amssymb}
\usepackage{graphicx}
\usepackage{subfig}
\usepackage{amsmath}
\usepackage{algorithm}
\usepackage{algorithmic}
\usepackage{float}
\usepackage{multirow}
\usepackage{scrextend}

\usepackage{tablefootnote}
\usepackage{multirow}
\usepackage{amsmath}

\journal{Knowledge-Based Systems}

\begin{document}

\begin{frontmatter}

\title{COVIDRead: A Large-scale Question Answering Dataset on COVID-19}

\author[mymainaddress1]{Tanik Saikh\corref{CorresAuthor}}
\cortext[CorresAuthor]{Corresponding author}
\ead{tanik4u@gmail.com}
\author[mymainaddress1]{Sovan Kumar Sahoo}
\author[mymainaddress1]{Asif Ekbal}
\author[mymainaddress2]{Pushpak Bhattacharyya}

 \address[mymainaddress1]{Department of Computer Science and Engineering \\Indian Institute of Technology Patna\\Bihta, Patna, India}
 \address[mymainaddress2]{Department of Computer Science and Engineering \\Indian Institute of Technology Bombay \\Mumbai, Maharashtra, India}

\begin{abstract}
During this pandemic situation, extracting any relevant information related to  COVID-19 will be immensely beneficial to the community at large. 
In this paper, we present a very important resource, 
\textit{COVIDRead}, a Stanford Question Answering Dataset (SQuAD) like dataset over more than 100k question-answer pairs. The dataset consists of Context-Answer-Question triples. Primarily the questions from the context are constructed in an automated way. After that, the system-generated questions are manually checked by humans annotators. This is a precious resource that could serve many purposes, ranging from common people queries regarding this very uncommon disease to managing articles by editors/associate editors of a journal. 
We establish several end-to-end neural network based baseline models that attain the lowest F1 of 32.03\% and the highest F1 of 37.19\%.  
To the best of our knowledge, we are the first to provide this kind of QA dataset in such a large volume on COVID-19. This dataset creates a new avenue of carrying out research on COVID-19 by providing a benchmark dataset and a baseline model. 
\end{abstract}


\end{frontmatter}


\section{Introduction}
\label{intro}
The Coronavirus diseases have affected human beings all over the world. This is a very new but infectious disease that are very unknown to the common men. So getting any kinds of information on this COVID-19 is of utmost need to all of us. Extracting such information is a very challenging problem. There is a pressing need that will serve the purpose of extracting information to any kinds of queries. A Question Answering (QA) system that would give answer to a query would serve this purpose. 
Recent advances in Artificial Intelligence (AI) have matured the task of field specific Question-Answering (QA). Lately, AI based deep learning techniques have shown its potential in many downstream Natural Language Processing (NLP) tasks. To build any end-to-end deep learning based QA system, dataset with a large volume is needed. This paper tries to bridge this particular gap. It offers a large volume of QA dataset to the research community to carry forward the research on COVID. In response to the COVID-19 pandemic, the White House and a coalition of leading research groups have prepared the COVID-19 Open Research Dataset (CORD-19) \cite{wang-etal-2020-cord}. CORD-19 is a resource of over 500k scholarly articles, including over 200k with full text, about COVID-19, SARS-CoV-2, and related coronaviruses. This freely available dataset is a valuable resource to the research community that propels in applying recent advances NLP and other AI based techniques (specially deep learning based approaches) to generate new insights in support of the ongoing fight against this infectious disease. There is an emerging importance for these approaches because of the rapid growth in the number of coronavirus related literature, making it difficult for the medical research community to keep up. \\
The ongoing pandemic of coronavirus disease in 2019
(COVID-19) was first reported in Wuhan, China in December 2019. The coronavirus disease is caused by severe acute respiratory syndrome coronavirus 2 (SARS CoV 2) and is primarily spread among people in proximity (within about 6 feet) most often via droplets produced by sneezing, coughing, talking. As the reports of the World Health Organization (WHO), no genuine licensed vaccines are yet available. Hence, the key public health strategies, such as surveillance, contract tracing, isolation and quarantine (wherever necessary) become the core methods to combat the deadly disease. \\
This kind of QA would help researchers, developers, editors/associate editors of a Journal and also to answer common men queries. The key attributes of the current work are summarized as below:
 \begin{itemize}
        \item \textit{This paper offers a dataset of QA. The dataset is having more than 100k instances of Context-Answer-Question triples. Out of which 40k questions are being human-annotated. This dataset is suitable for training data sensitive various end-to-end deep neural network models.}
        \item \textit{A set of 1000 question-answer pairs are prepared manually by human experts. Human made questions after reading the articles and put his/her answers for that particular question. We provide answers for each question and its' supporting text. So we prepare 1k such example pairs having abstract-question-answer triples. This can be used as a gold standard test for future research. }
        \item \textit{We propose several baseline models based on BERT, BioBERT, and ClinicalBERT.}
 \end{itemize}

\section{Related Work}
The paper by  
\cite{moller-etal-2020-covid} first presented dataset for QA on COVID named as \textit{COVID-QA}. They made use of the CORD-19 scientific articles. They offered biomedical experts annotated 2019 question-answer pairs related to COVID-19 by considering 147 scientific articles. They provided RoBERT based baseline, fine-tuned on SQuAD along with model fine-tuned on the created \textit{COVID-QA} dataset. The model obtained the performance in terms of \textit{Exact Match} and \textit{F1} as 25.9\% and 59.53\%, respectively. \\
The work defined in 
\cite{su-etal-2020-caire} offered CAiRE-COVID system. The system consists of three modules, \textit{viz.} \textit{Document Retriever}, \textit{Relevant Snippet Selector}, and \textit{Multi-Document Summarizer}. The first two modules perform an open-domain QA, whereas the last one is responsible for generating concise summary (abstractive and/or extractive) of the top-ranked retrieved relevant paragraphs. To better generalise the system to unseen questions, the system is trained in a multi-task learning scheme on six datasets, namely \textit{SQuAD}, \textit{NewsQA}, \textit{TriviaQA}, \textit{SearchQA}, \textit{HotpotQA}, \textit{Natural Question}. The system is evaluated on Covid-QA dataset. The paper \cite{otegi-etal-2020-automatic} proposed a system that combines an Information Retrieval module and a reading comprehension module that finds the answers in the retrieved passages. \\

The paper described in \cite{DBLP:journals/corr/abs-2006-15830} provided a real-time QA system, namely \textit{covidASK}. This also offered a small amount of dataset on Covid-19. The \textit{covidASK} system has borrowed many techniques from Open-domain QA. \\
The task defined in \cite{tang2020rapidly} presents \textit{CovidQA} that comprises 124 question-article pairs. This is a very small in size. This dataset is not suitable for running any deep neural network models. But it can be helpful for evaluating the zero-shot or transfer capabilities of the existing models on COVID-19 domain.\\
The paper of \cite{reddy2020end} 
shows an effective end-to-end method for zero-shot adaptation for an open domain QA system to a target domain, COVID-19 in this case. They also proposed a novel example generation model that can produce synthetic training examples for both information retrieval and machine reading comprehension.\\
\cite{DBLP:journals/corr/abs-2006-09595} proposed CO-Search, that is a semantic, multi-stage search engine designed to handle complex queries over the COVID-19 literature. CO-Search comprises a hybrid semantic-keyword retriever and a re-ranker. The retriever (Siamese-BERT based model) outputs a list of relevant documents given a query. The re-ranker orders them by relevance that also assign a relevance score to each document. However, they test the efficacy of the system on the combination of CORD-19 corpus and the TREC-COVID competition’s evaluation dataset. \\    
The task that we tackle here is based on QA over COVID-19. The dataset is human-annotated and large in size. We provide a rigorous experimentation and detailed error analysis. We believe that our dataset is the largest one among the datasets published so far. This kind of dataset is suitable for developing domain-specific end-to-end automated deep neural models.
 
\section{Dataset Creation}
We create \textit{COVIDRead}, a dataset for supporting text grounded extractive QA system on COVID-19. This section describes our dataset creation procedure including data collection, data pre-processing etc. In light of COVID-19 pandemic, the white house and coalition of leading research groups have prepared the COVID-19 Open Research Dataset (CORD-19) \cite{DBLP:journals/corr/abs-2004-10706}. CORD-19 is a collection of various scholarly articles, about COVID-19, SARS-CoV-2, and related corona viruses. These articles were collected from the sources like PMC, Elsevier, WHO, Medrxiv, Biorxiv etc. We collect the articles available in the Kaggle Portal \footnote{https://www.kaggle.com/allen-institute-for-ai/CORD-19-research-challenge}. We collect 51,078 (as these many were available at the time of starting our work) such articles. Among them, 8726 such articles were there whose abstracts are missing. We consider the remaining (i.e., 42,359) articles for the experiment purpose. These articles are in Comma-Separated Values (.csv) format each column having components like abstract, introduction etc. We extract abstracts for each article. Now we are having a collection of 42,359 abstracts. We split each abstract into sentences using NLTK sentence splitter \footnote{https://www.nltk.org/api/nltk.tokenize.html}. We pass each sentence of a particular abstract through a constituency parser. We foster Stanford CoreNLP \footnote{https://stanfordnlp.github.io/CoreNLP/} for this purpose. Constituency parser essentially breaks each sentence contained in the abstract into noun and verb phrases. We take noun phrases into account. These noun phrases become the plausible answers for that particular abstract \footnote{We use the terms Abstract, Document and Paragraph interchangeably throughout this paper.}. 

The studies like \cite{rajpurkar-etal-2016-squad,trischler-etal-2017-newsqa} have suggested that noun phrases of a particular document are the plausible answer of that particular document. We borrow the concept from there and utilized in our task. These answers are paired with its' corresponding document. Now, we are having document and answer pairs. These pairs are further passed through a Question Generator (QG) model. The QG model produces questions conditioned on the given abstract and answer pairs. Now we are having \textit{Context-Answer-Question} triples. In this way, we generate over 100k such pairs. Keeping the quality of the generated questions in mind, we employ 3 annotators having expertise in medical domain with very good knowledge in the English language. The annotators manually checked the generated questions and also edited. This way we have created over 100k data-points out of which 18k are human annotated. We coin this newly created dataset as \textit{\textbf{CovidRead}} \footnote{We will make the dataset publicly available after acceptance of this paper.}.

We split the whole dataset into training, development, and testing sets, as shown in Table \ref{tab:data-distribution}.
\begin{table}[h!]
\centering
\begin{tabular}{|l|l|l|l|}
\hline
 & Training & Development & Testing \\ \hline
CovidRead & 82415 & 10000 & 10000 \\ \hline
\end{tabular}
\caption{The standard distribution of CovidRead into Training, Development, and Testing sets}
\label{tab:data-distribution}
\end{table}

 \begin{figure*}[!ht]
\centering
  \includegraphics[width=7cm,height=12cm]{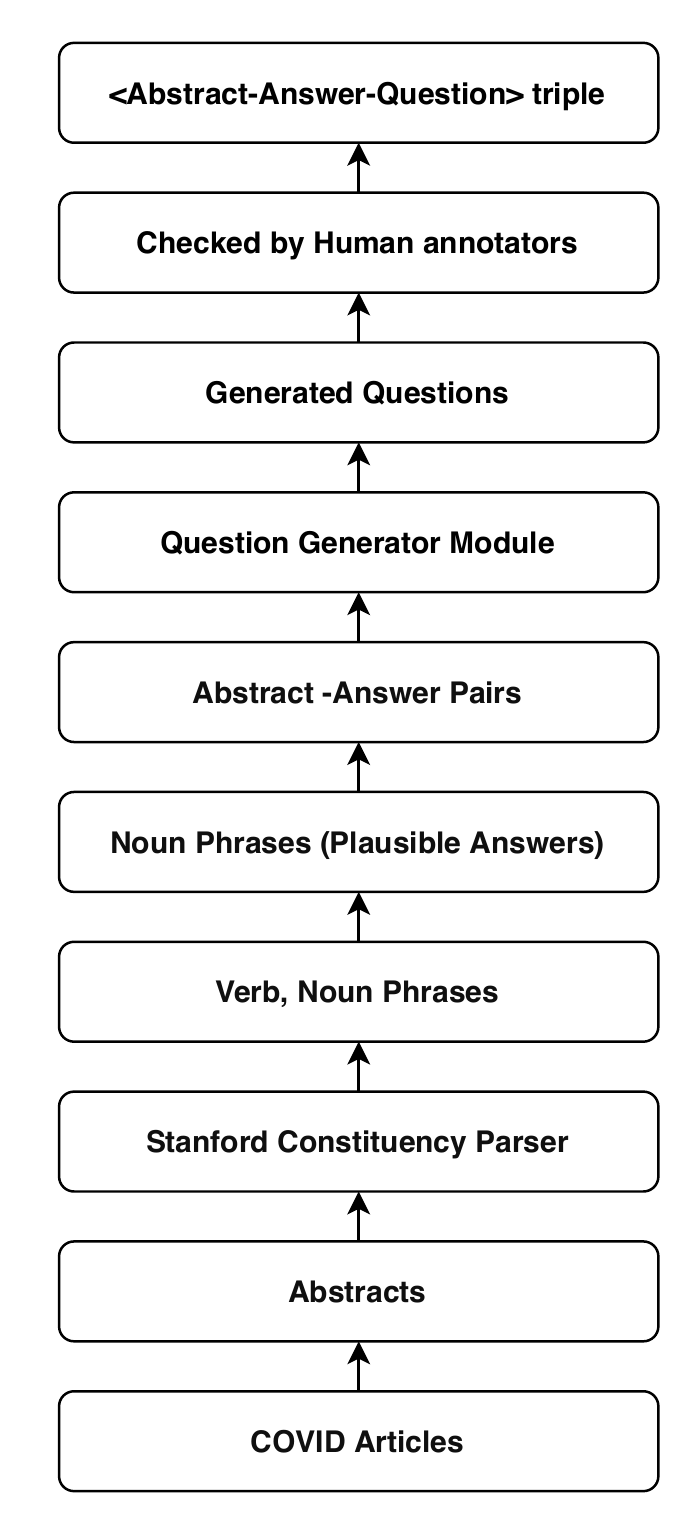}
  \caption{The Proposed System for Synthetic data
 generation from COVID related Research Articles}
  \label{fig}
\end{figure*}
\subsection{Evaluation of Generated Questions}
Keeping the quality of the generated questions in mind we perform various evaluation schemes i.e. automatic as well as manual. Evaluating the outcomes of any Natural Language Generation (NLG) systems is much more complex compared to that of conventional NLP tasks (e.g., sentiment analysis, semantic parsing) where only label matching is required. For NLG, there are no such scopes. We apply the metrics that are widely used for Machine Translation and Summarization (i.e., BLEU \cite{bleu}, METEOR \cite{banerjee-lavie-2005-meteor}, and ROUGE \cite{lin-2004-rouge} etc.) tasks. We apply these metrics on a sample of 2000 output examples for evaluation. These evaluations metrics' outputs are shown in the Table \ref{tab:eval-results}. The BLEU, METEOR, and ROUGH are the n-gram based metrics, where n-gram lexical matching is performed. We also apply Consensus-based image description evaluation (CIDEr) metric, which is a popular one for evaluating various tasks in computer vision. Apart from these metrics, we also perform a human evaluation. For this we employ human annotators. Additionally, we propose a novel metric that could be a potential metric for any kinds of NLG tasks. The approach is based on logic (i.e., Entailment \cite{DBLP:conf/mlcw/DaganGM05, maccartney2009natural}). Textual Entailment (TE) refers to the task of determining whether sentence A is the logical consequence of another sentence B or not. We can argue that this would the most appropriate metric to judge whether two pieces of sentences are similar or not. With this motivation we determine the entailment relation between machine-generated questions and reference questions by applying one of the state-of-the-art entailment models that is equipped with external knowledge \cite{chen-etal-2018-neural}. The model is trained with the combination of SNLI and Multi-NLI \cite{N18-1101}. We evaluate the trained NLI system by giving the inputs (i.e., system generated and human-generated questions as premise and hypothesis respectively) to predict the entailment relation between the system generated and human-generated questions. We asked the annotators to provide inference labels (i.e., Entailment, Contradiction, and Neutral) to each reference question given the corresponding system-generated questions. We compare these labels with the system's predicted labels. Further, we compute the accuracy and got around 72\%. Entailment relation between these two questions indicates that generated questions are logically accurate and correct, i.e., much closer to the question as humans generally used to ask. This Entailment based metric could be a promising evaluation metric for any NLG tasks. 
\begin{table}[h!]
\centering
\begin{tabular}{|l|l|l|l|}
\hline
BLEU & METEOR & ROUGE & CIDEr \\ \hline
0.18 & 0.12 & 0.12 & 0.345 \\ \hline
\end{tabular}
\caption{Evaluation results on NLG metrics}
\label{tab:eval-results}
\end{table}
\subsection{Annotation Guidelines}
We employ three annotators to check the quality of the system generated questions for a period of more than one and half year i.e. from the inception of COVID-19 pandemic. The annotators are of post-graduate level in Bio Science with an age group of 35-50 and having good expertise in English. We generally do hypothesis testing while doing any kind of evaluation. 
We consider the following hypotheses for the generated questions i.) Are the generated questions appreciable by a human? and ii.) Are the generated questions readable and comprehensible easily and also grammatically correct? Keeping these parameters in mind, we instructed the annotators as follows:
\begin{itemize}
  \item The generated questions should be grammatically correct, and that should be spelled correctly; no conjunction, proper punctuation should be there; questions and proper nouns should begin with a capital letter.
  \item The question should be understood by anyone, even to someone who is totally unaware of the context.
  \item For all the factoid questions, the answer should be unique and factual, and for others types of questions the answers should be descriptive as \textit{COVIDRead} is a mixture of factoid and non-factoid questions.
\end{itemize}

We randomly chose 500 samples. To judge the naturalness (i.e., verify how the generated questions are grammatically correct and how much fluent), we employ two annotators. They were asked to give scores (between 0-4) to each such question based on the above two parameters. We compute the inter-annotator agreement ratio in terms of the kappa coefficient \cite{fleiss1971measuring}. It was found to be 0.83, which is considered to be good agreement as per \cite{landis1977measurement}.
\subsection{Dataset Analysis}
The \textit{CovidRead} contains over 100k synthetic data points, i.e., triples of Context-Answer-Question. 
We manually check 40K such instances. We make a rigorous analysis of these data points. The following Table \ref{tab:stat-data} shows the average length of Context, Answer, and Question. All the lengths are larger compared other datasets in this line. 
\begin{table}[h!]
\centering
\begin{tabular}{|l|l|l|l|}
\hline
 & \multicolumn{1}{c|}{\textbf{Context}} & \multicolumn{1}{c|}{\textbf{Question}} & \multicolumn{1}{c|}{\textbf{Answer}} \\ \hline
\multicolumn{1}{|c|}{\textbf{Average Length}} & 1616.84 & 52.31  & 28.46\\ \hline
\end{tabular}
\caption{Average length of Context, Question, and Answer in the CovidRead dataset}
\label{tab:stat-data}
\end{table}

To understand the properties of the \textit{CovidRead}, we analyse the Questions and Answers in the Training/Development set. We also categories our dataset into factoid and non-factoid questions. \\
\textbf{Diversity in Answer:} From the Table \ref{tab:stat-data}, it is shown that the average length of the answers is pretty high. So these should not be the answers of only factoid questions. 
These could be the answer of non-factoid questions too. We assumed NPs of a passage are the plausible answers for that passage. So sometimes constituency parser produces incomplete noun phrases, that too become our answers. In this way we obtain incomplete answers sometimes, such as : \textit{"CT equipment installation and environment"} turned into \textit{"installation and environment"}. This could be fault of the constituency parser used.\\

\begin{figure*}[!ht]
\centering
  \includegraphics[width=15cm,height=10cm]{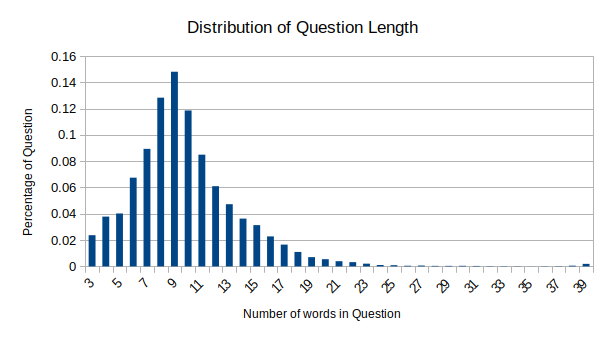}
  \caption{Percentage of questions with different word length}
  \label{fig-bar}
\end{figure*}

As we assumed that the NPs are the answers for a particular document, our answers' types are noun phrases. \\
\textbf{Diversity in Questions:} Over the time, QA systems have matured enough that can answer factoid questions quite comfortably but they are far from giving answer to non-factoid queries. 
For an accurate QA system that would produce accurate answer to queries, understanding of questions is important that requires to understand the type of questions. We compute percentage of each question for each type of Questions. Quantitatively, the percentage of questions for different types is shown in Table \ref{tab:Question-stat}. Among them, the maximum number of questions are of \textit{What} type. As the Table \ref{tab:Question-stat} shows that the other types are categorized into \textit{"When", "Where", "Who"} types etc. in descending order. To better understand the questions contained in our dataset we compute the distribution of question length and percentage of questions that are confined within a particular range of length that is shown in Figure \ref{fig-bar}. 

\begin{figure*}[h!]
\centering
  \includegraphics[width=15cm,height=10cm]{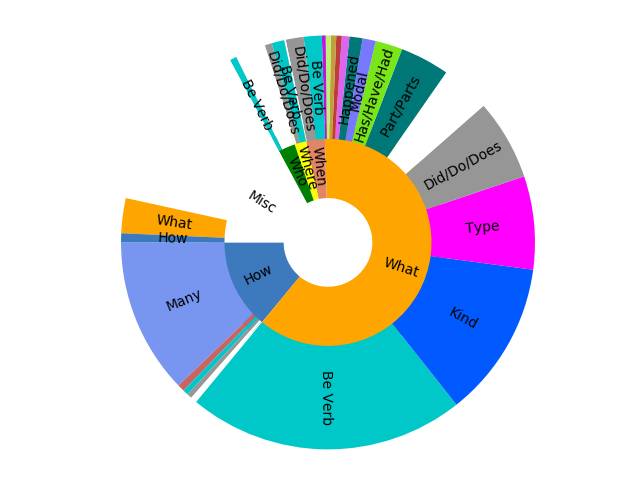}
  \caption{Distribution of questions by their first two words for a random sample of 40K questions. The arc length is proportional to the number of questions containing the word.}
  \label{fig-donate}
\end{figure*}

We see from the Figure \ref{fig-bar} that most questions range from seven to ten words that is quite logical. We discard the questions along with its context and answer from the dataset, having number of words less than 3. These types of questions could be occurred in community question answering scenario or QA on ChatBot. As we are not in such scenarios, we have omitted those questions. To get better view of our questions classification, we plot another chart, namely a multi-level donate chart. Figure \ref{fig-donate} shows such chart. The distribution of questions is based on the first two words of the questions. There exists a surprising variety, like what followed by be verb (is, am, are, was, were etc) forming the maximum number of questions. Then \textit{What} followed by \textit{kind/type} are the second most questions. In \textit{How} categories \textit{How many} type is the most number of questions. In \textit{Misc} category other types of questions including those that are not starting with any \textit{wh} words, it may contain at the middle and end of the question. Understanding questions is one of the key topics in any QA problem.\\

\begin{figure*}[h!]
\centering
  \includegraphics[width=15cm,height=10cm]{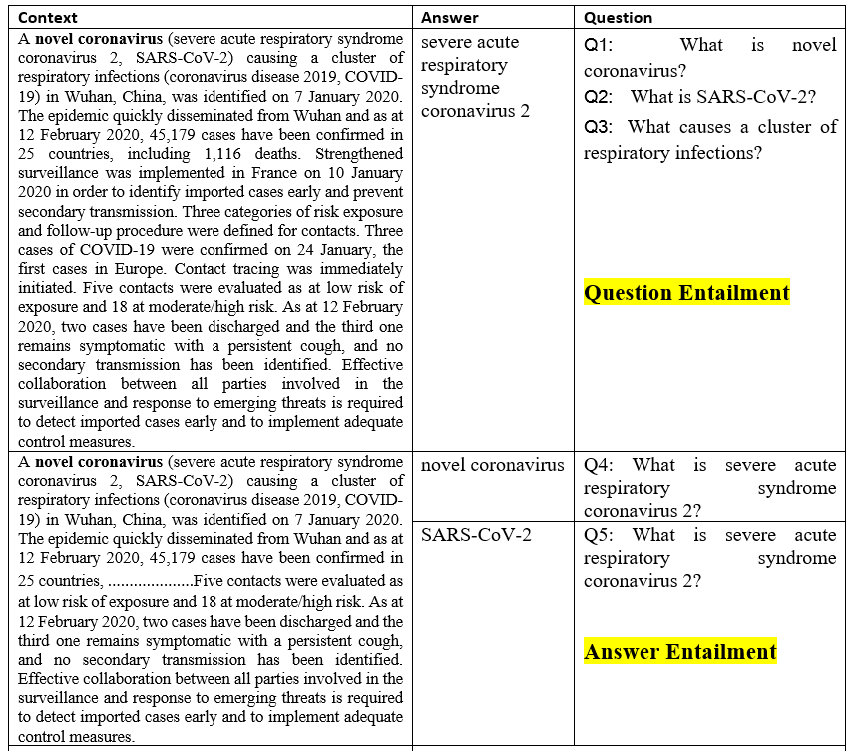}
  \caption{Example of data set having Class - I questions. Table also shows the examples of Question Entailment and Answer Entailment.}
  \label{fig:goodExam}
\end{figure*}

\begin{figure*}[!ht]
\centering
  \includegraphics[width=15cm,height=10cm]{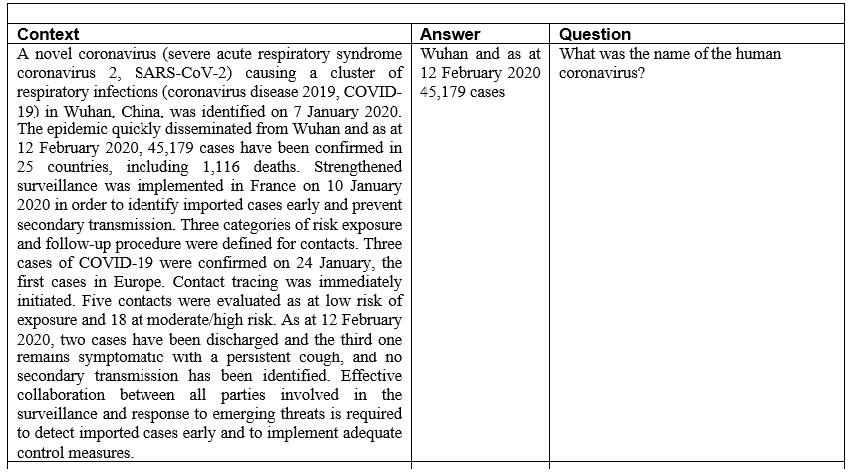}
  \caption{Example of Data set having Class - III questions.}
  \label{fig:badExam}
\end{figure*}

\begin{figure*}[!ht]
\centering
  \includegraphics[width=15cm,height=10cm]{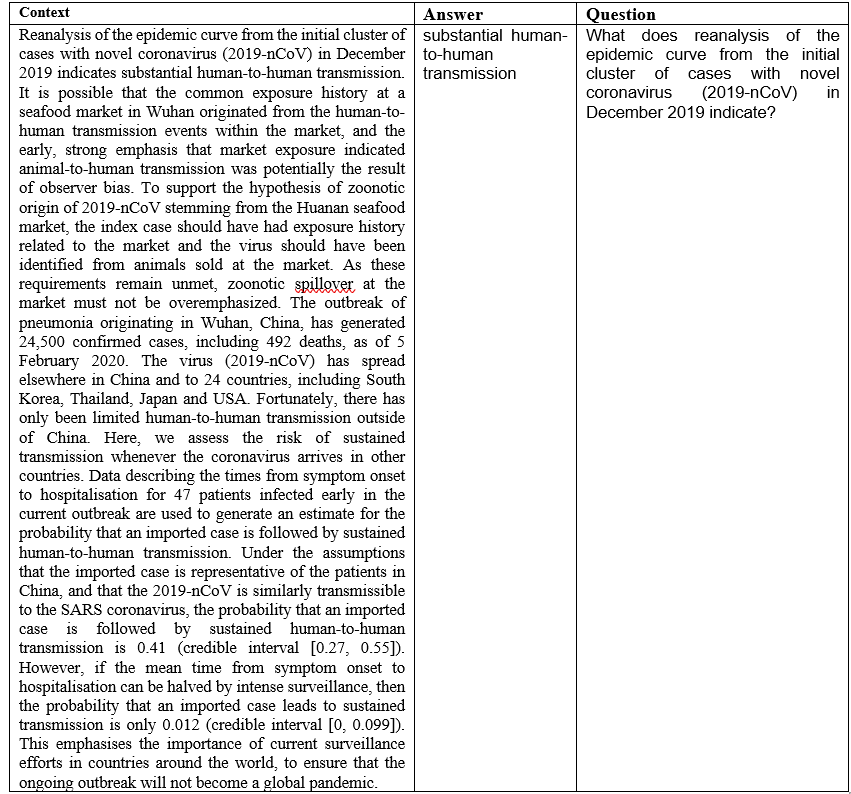}
  \caption{Example of Data set having Class - II questions.}
  \label{fig:avgExam}
\end{figure*}

We also perform a rigorous analysis of 1000 sample questions and categories them into \textit{Class-I, Class-II, and Class-III}. Class-I defines the set of questions that are easily understandable, grammatically correct, contextual and also  understandable even to those who are unaware of the context at all. It is observed that 35\% of the questions lay in this category. 
Questions that are understandable to some extent with some minor mistakes in grammar fall into Class-II category. There are 40\% such questions. 
Remaining 25\% of the questions could be categorized into \textit{Class-III} category, i.e. there are incomplete questions, not understandable and the grammar is also poor. The Figures \ref{fig:goodExam}, \ref{fig:avgExam} and \ref{fig:badExam} show the examples of Class-I, Class-II and Class-III respectively. \\
\begin{table}[h!]
\centering
\begin{tabular}{|l|l|}
\hline
\textbf{Question Type} & \textbf{\% of Questions} \\ \hline
What & 78.76 \\ \hline
When & 3.14 \\ \hline
Where & 2 \\ \hline
Who & 37.44 \\ \hline
Whom & 0.11 \\ \hline
Which & 0.16 \\ \hline
Whose & 0.03 \\ \hline
Why & 0.72 \\ \hline
How & 9.73 \\ \hline
How many & 7.7 \\ \hline
Is/Are & 0.48 \\ \hline
\end{tabular}
\caption{Percentage of questions for each type of question }
\label{tab:Question-stat}
\end{table}
\textbf{Question Entailment and Answer Entailment:}  \textit{Question Entailment} and \textit{Answer Entailment} drive its definition from the basic definition of Text Entailment \cite{dagan2005pascal}. \cite{groenendijk1984studies} defines Question Entailment between two Questions ($Q_1$ and $Q_2$) if every proposition giving an answer to $Q_1$ is also giving an answer to $Q_2$. \cite{roberts1996information} considered the first question $Q_1$ as a superquestion $Q_2$ as a subquestion, if we answers to subquestions then we have the answer to superquestions. From these two definitions we can say, Question Entailment is defined as question A entails question B if every answer to B is also a complete or partial answer to A \cite{abacha2016recognizing}. Conversely, we can introduce the concept of Answer Entailment as, if two answers say: Answer A and Answer B are the answers of a same question and/or its paraphrased question, then it could be termed as Answer Entailment. In the Figure \ref{fig:goodExam}, we show the example of these two types of entailment. The aim of recognizing question entailment (RQE) is to find answer to an incoming question by mining entailed questions with its associated answer. This can therefore be defined as: \\
a question A entails a question B if every answer to B is also a complete or partial answer to A.  
The figure \ref{fig:goodExam} shows the three questions Q1, Q2 and Q3 pointing to the same answer. So this is the example of Question Entailment. On the contrary, Answer Entailment could be termed as follows:\\
If one question has two/multiple answers (Answer A and answer B), then this scenario could be termed as the answer entailment. Refer to the Figure \ref{fig:goodExam}, where the question 4 and question 5 have the same answers. This is therefore an example of Answer Entailment. This Answer Entailment could help in Question Answering, specially in handling contradictory questions. We identify such kinds of Question Entailment and Answer Entailment examples in our \textit{CovidRead} dataset.
\\

As said, we have extracted these two type of examples from our dataset. This provides a new direction of improving QA system. One of the studies like \cite{DBLP:journals/corr/abs-1901-08079} has showed the potential of Question Entailment in improvement of QA system. There are no such studies that have shown Answer Entailment helping QA.\\
\textbf{Reasoning required to answer questions:} To get a better understanding of the reasoning required to answer the questions, we picked up examples randomly. We analyse those examples, i.e., what are required to give the answer to the questions, and like SQuAD we manually label the examples with the categories shown in Table \ref{tab:reasoning}. We have categorised them into a few categories. 

In the Table \ref{tab:reasoning}, in the first row to answer the question system has to take care of co-reference resolution. To answer the question contained in the second row, QA system has to efficiently handle the negations. Handling negation in any context is one of the toughest problem in NLP. The contexts are full of Named Entities (NEs) (specially contain medical NEs), so maximum questions are formed whose answers are NEs. To properly answer the questions contained in the third row, system has to efficiently disambiguate the appropriate NE among multiple NEs 
contained in the supporting text. 

 \begin{table}[h!]
 \centering
 \begin{tabular}{|l|l|l|}
 \hline
 \textbf{Reasoning} & \textbf{Context} & \textbf{Question} \\ \hline
 \begin{tabular}[c]{@{}l@{}}Over multiple \\ sentences\end{tabular} & \begin{tabular}[c]{@{}l@{}}On December 31, 2019, several cases \\ of pneumonia of unknown etiology \\ have been reported in Wuhan, Hubei \\ province, China {[}1,2,3{]}. On January 7, \\ 2020, Chinese health authorities confirmed \\ that these cases were associated with a novel\\ coronavirus, which was subsequently named \\ 2019­nCoV by WHO {[}4{]}.\end{tabular} & \begin{tabular}[c]{@{}l@{}}Which cases were \\ associated with a novel \\ coronavirus that was \\ confirmed by Chinese \\ health authorities on \\ January 7, 2020?\end{tabular} \\ \hline
 \begin{tabular}[c]{@{}l@{}}Negation \\ Handling\end{tabular} & \begin{tabular}[c]{@{}l@{}}Our integrative study not only presents \\ a systematic examination of \\ SARS-CoV-2-induced perturbation of\\ host targets and cellular networks to reflect\\ disease etiology, but also reveals insights into \\ the mechanisms by which SARS-CoV-2 \\ triggers cytokine storms and represents a \\ powerful resource in the quest for therapeutic \\ intervention.\end{tabular} & \begin{tabular}[c]{@{}l@{}}What does this \\ investigation \\ present?\end{tabular} \\ \hline
 \begin{tabular}[c]{@{}l@{}}Named Entity \\ Recognition\end{tabular} & \begin{tabular}[c]{@{}l@{}}A novel coronavirus (severe acute respiratory\\ syndrome coronavirus 2,SARS-CoV-2) \\ causing a cluster of respiratory infections \\ (coronavirus disease 2019, COVID-19) in \\ Wuhan, China, was identified on 7 January 2020.\end{tabular} & \begin{tabular}[c]{@{}l@{}}Q1: What is SARS-CoV-2?\\ Q2: What is novel \\ coronavirus?\end{tabular} \\ \hline
  &  &  \\ \hline
 \end{tabular}
 \caption{We manually labeled $\sim$200 examples into one or more of the above categories.}
 \label{tab:reasoning}
 \end{table}

From Table \ref{tab:reasoning} it is obvious that some percentage of questions in \textit{CovidRead} are not straightforward to answer, whereas many are so easy to answer.

\section{Methods}
\begin{figure}
     \centering
    \includegraphics[width=80mm,scale=0.8]{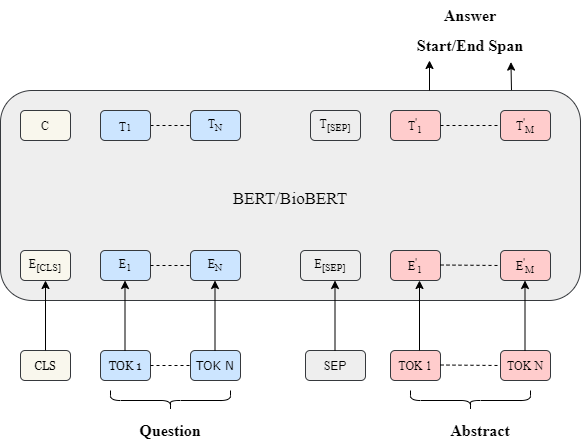}
    \caption{A BERT/BioBERT based QA Model. Image courtesy: (\cite{devlin-etal-2019-bert}; https://blog.scaleway.com/2019/understanding-text-with-bert/) with minimal modifications}
     \label{fig:bert}
 \end{figure}

We propose different transformer based methods for QA. Transformer based models have shown tremendous progress in various downstream tasks in NLP. Actually, we make use of bi-directional version of this network. The following subsections are showing the different models used in this paper. \\
\textit{\textbf{BERT:}} Any QA system requires understanding of the underlying language. Recently transformer architecture based models have shown this kind of capability of capturing language understandability, specifically the bi-directional version of this. Keeping this point in mind we apply this transformer based model into our COVID QA task. This is our simple baseline model. The diagram is shown in the Figure \ref{fig:bert}.\\
\textit{\textbf{BioBERT:}} Our second model is BioBERT \cite{10.1093/bioinformatics/btz682} based. This is pre-trained biomedical language representation model. It is initialized with BERT's pre-trained parameters. After pre-training, the model is further fine-tuned on PubMed abstract and PubMed Central full-text articles. We apply the best version of this model. The motivation is as we are dealing with the vocabularies that contain bio-medical texts, we make use of this model. The architecture is same as the previous one with slide modification in the training data. \\
\textit{\textbf{ClinicalBERT:}} Our third model is another version of this language model, i.e. ClinicalBERT \cite{alsentzer2019publicly}. This model is initialised from BioBERT v1.0, and then pre-trained on more than 2 million clinical notes contained in the MIMIC-III v1.4 database of patient notes \cite{johnson2016mimic}. We make use of the best variant of this of Clinical BERT that has 108M parameters.

\section{Experiments, Results and Discussions}
We run the baseline and the proposed models on our created dataset. We perform exhaustive experiments with these models by different combinations of the datasets. We perform six experiments.  \\
\textbf{BERT:} There is a recent style of knowledge learning and utilizing vibrates the NLP field with huge success authorized by the pre-trained language model (i.e., BERT) and its variants. This model is based on vanilla BERT. We follow the architecture provided in the Figure \ref{fig:bert}. We have 100,000 examples (synthetic) of Abstract-Question-Answer triples. Out of which nearly 40K questions are manually checked. We train the model on 39178 such instances, and tested with 1000 instances. The result obtained is shown in the Figure. From the Figure \ref{fig:bert}, it is clear that the system takes concatenation of Context and Question (context and question separated by [SEP] token that starts with [CLS] token) as input. The system produces span of texts as output, that is being the predicted answer.  
There are two classifiers on top of this model. One classifier (with a start vector $S$) predicts start index and the second one (with end vector $H$) yields end index of the answer. The probability of token being start of an answer is computed by the dot product $T_i$ and a start vector $S$. Then there is a soft-max over all the tokens in the abstract. 
\begin{equation}
P_i  = \frac{e^{ST_i}}{\sum_j e^{ST_i}} 
\end{equation}
In the similar way, the end index of the answer span is calculated. Candidate span score from i to j is defined as $S.T_i+E.T_j$. The span having maximum score (where $j>=i$); is considered as the predicted answer. We take the implements of the models available in huggingface. 
\\
\textbf{BioBERT} The second experiment is performed using BioBERT based model. This is basically BERT model enriched with biomedical knowledge via pre-training over biomedical corpora like PubMed. Keeping the same data distribution we perform this experiment too. Evaluation results of this model are shown in the Table \ref{tab:results}. \\
\textbf{ClinicalBERT} Our third experiment is performed using the clinical BERT model. In this experiment the data distribution is also the same. Results obtained by this model are shown in Table \ref{tab:results}.\\
\textbf{BioBERT-I} This experiment is based on the bioBERT model with some modifications. We named this as BioBERT-I. We apply a SQuAD pre-trained QA model and that further find-tuned on the Gold Standard dataset. We test this model with our test set of 1000 instances. The results produced by this model are shown in Table \ref{tab:results}. \\
\textbf{BioBERT-II} This model is trained on the whole dataset i.e. mixture of human annotated and synthetic data. We called this model as BioBERT-II. The size of such dataset is over 100k instances. We further test this trained model with the gold standard test dataset having 1000 instances.\\
\textbf{BioBERT-III Synthetic-Train-fineTuned:} This experiment is same the previous one but slightly different from the previous one. Here, we perform the training of this model with the same dataset as the previous one, then fine-tune and then trained. We further text the trained  model with the gold standard dataset. The results obtained are shown in Table \ref{tab:results}. \\

We rely on two standard evaluation metrics (i.e., Exact Match and F1) that are widely used for evaluating Span-of-Texts based extractive QA systems \cite{rajpurkar-etal-2016-squad,lai-etal-2017-race}. The metrics avoid punctuation and articles (e.g., a, an, the etc.). We have only one reference answer for testing. Using this we compute \textit{Exact Match} and \textit{F1} scores. These two are the standard metrics for evaluating Span-of-Texts based QA Systems, and widely used in various notable QA Systems \cite{rajpurkar2016squad,trischler-etal-2017-newsqa,JoshiTriviaQA2017,lai-etal-2017-race}. \\ 
\textbf{Exact Match: } This metric computes the number of matching predicted answers with the ground truth answers character by character. \\
\textbf{F1 score: } This is a macro-averaged F1 score. It converts the predicted and ground-truth answers as bag-of-words. Then calculate the average overlap between the predicted and ground-truth answers. Further, computes their F1, and then take the average over all the instances. \\
The results obtained are shown in Table \ref{tab:results}. From the Table, we can say that all the proposed models perform better compared to the baseline in terms of exact match. 
The vanilla BERT model is fine-tuned on a large dataset that performs better compared to the baseline. In these experiments, we deal with biomedical scientific texts, so our next model utilises the BERT model that is trained on biomedical scientific texts (i.e., BioBERT). This model performs better compared to the vanilla BERT model. Our third proposed model is based on BERT that is trained on Clinical Bio-medical texts. It perform better compared to BERT and also Bio-BERT based model. 
Our next model is based on Bio-BERT that is pre-trained on SQuAD and further trained and tested on the gold data. This model is performed inferior compared to the last three models. As SQuAD is full of factoid questions, and our dataset is a mixture of factoid and non-factoid questions, that may be the cause of poor performance of this model.
The next model is also based on Bio-BERT but train on whole 100k examples and tested on the gold test data. The model yields superior results compared to the previous models. As training and testing is perform on the same domain dataset, this is one of the reasons of good performance. The last model is same as the previous one but fine tuned on the gold dataset. This one performs the best among all the proposed models in terms of both the evaluation schemes.


\begin{table}[]
\centering
\begin{tabular}{|c|c|c|}
\hline
\multirow{1}{*}{Models} & \multicolumn{2}{c|}{Results (\%)} \\ \cline{2-3} \hline
 & Exact Match & F1 \\ \hline
\multicolumn{3}{|c|}{Proposed} \\ \hline
BioBERT-III & 28.65 & 37.19 \\ \hline
BioBERT-II & 25.2 & 33.95 \\ \hline
BioBERT-I & 22.25 & 29.93 \\ \hline
ClinicalBERT & 25 & 34.09 \\ \hline
BioBERT & 24.95 & 34.37 \\ \hline
BERT & 23.2 & 32.03 \\ \hline
\multicolumn{3}{|c|}{State-of-the-art} \\ \hline
COVIDQA Model & 25.90 & 59.53 \\ \hline
\multicolumn{3}{|c|}{Baseline} \\ \hline
COVIDQA Model & 21.84 & 49.43 \\ \hline
\end{tabular}
\caption{Proposed methods' results and comparison with baseline model.}
\label{tab:results}
\end{table}

\subsection{Error Analysis}
We perform qualitative and quantitative error analysis. We rigorously compare the actual and predicted output. The following are our observations:
\begin{itemize}
  \item Most of BERT-like models have limitations of max input of 512 tokens, but in our case, some contexts are having more than 2000 tokens. BERT models trim the remaining tokens, resulting information lost in the context. It would be one of the reasons for low performance. 
  \item The system is able to predict the answer of the questions, where there is a token overlap between question and answer containing the sentence. 
  \item Sometimes the system talking the whole phrase before the generating question sentence. 
  \item Sometimes the system taking the answer containing full phrase instead of the exact answer, as a result the system yields better F1 compared to exact match. 
  \item System unable to predict numerical type answer. 
  \item Abstracts are confined with 260-300 words, which is considered a long document compared to the document in SQuAD. Such long documents make the QA task difficult.
\end{itemize}

\subsection{Observation and main message}
We offer QA dataset having Context-Question-Answer examples triples over 100K 
from research articles to the research community on COVID-19. Out of 100K examples, 40K questions are human-annotated, the remaining are synthetic. We hope this novel dataset can serve as a precious resource for research and evaluation of machine reading comprehension models (specifically, QA and QG models) on research articles. 
Our test dataset is paraphrased. Testing a QA model with this parapharsed enriched data would test the robustness of a QA systems. So our dataset would serve this purpose also.  
\section{Robustness of QA model}
A question could be asked in many ways. We human being generally follow this trend and used to answer accordingly. To make an automatic QA system to act as human, machine has to be trained in this manner. For this we paraphrased our test set's each question upto 3 questions. We make use of a pre-trained paraphrased model \cite{DBLP:journals/corr/abs-1910-10683}
for this purpose. 
Then we augment our test set with these paraphrased questions. We test our best performing system again on this augmented test set, and observe that the results get decreased as shown in Table \ref{tab:paraphrase}. So to test the robustness of a QA system, our paraphrased enriched corpus will be useful. 
\begin{table}[]
\centering
\begin{tabular}{|l|l|l|l|}
\hline
\textbf{Data} & \textbf{Model} & \textbf{Exact Match} & F1 \\ \hline
\begin{tabular}[c]{@{}l@{}}Trained and tested on \\ paraphrased enriched corpus\end{tabular} & BioBERT & 20.77 & 28.86 \\ \hline
\begin{tabular}[c]{@{}l@{}}Trained and tested on \\ without paraphrased enriched corpus\end{tabular} & BioBERT & \textbf{29.91} & \textbf{36.88} \\ \hline
\end{tabular}
\caption{BioBERT based model tested on i.) paraphrased enriched corpus and ii.) without paraphrased enriched corpus}
\label{tab:paraphrase}
\end{table}

\section{Conclusion and Future Work}
In this paper, we have presented \textit{CovidRead}, a new dataset for benchmark evaluation of methods for QA/MRC on Covid-19. 
The dataset is created in a semi-automatic way. 
The dataset consists of Context-Question-Answer triples over 100k, out of which 40K are human-annotated. Researchers and editors/associate editors would be benefited out of this kind of models and the dataset. We have proposed three approaches that are based on \textit{viz.} (i.) Vanilla BERT (ii.) Bio-medical BERT (i.e., BioBERT) and (iii.) Clinical BERT. Our proposed models are competitive compared to the existing state-of-the-art models. Our future works would include:
\begin{enumerate}
    \item Considering the full-text of manuscripts instead of only abstracts. As  users' all questions could not be found in the abstract only.   
    \item Implementation of the Bi-directional Attention flow (Bi-DAF) model by incorporating multi-hop attention. 
    \item Enrichment of this dataset up to 500K in size.
    \item Multi-hop version of this CovidRead (like HotpotQA) dataset.
    \item Building a Visual Question Answering (VQA) datasets and models on the images available in the covid-19 literature. 
    \item Model based on Generative Pre-trained Transformer (GPT)-3. 
\end{enumerate}
\section{Acknowledgement}
On behalf of all the co-authors Mr. Tanik Saikh would like to acknowledge "Elsevier Centre of Excellence for Natural Language Processing" at Indian Institute of Technology Patna for supporting  research work carried out in this paper.

\bibliographystyle{elsarticle-num}
\bibliography{absa}

\end{document}